\documentclass{edm_article}
\usepackage{natbib}
\usepackage{amsmath,amssymb,bm}
\usepackage{booktabs}
\usepackage{url}
\newcommand{\ord}{\prec}
\newcommand{\figref}[1]{Fig.~\ref{#1}}
\newcommand{\tabref}[1]{Tab.~\ref{#1}}

\begin{document}

\title{Predicting Disagreement with Human Raters in LLM-as-a-Judge Difficulty Assessment without Using Generation-Time Probability Signals}

\numberofauthors{1}
\author{
Yo Ehara\\
       \affaddr{Tokyo Gakugei University}\\
       \email{ehara@u-gakugei.ac.jp}
}

\maketitle
\begin{abstract}
Automatic generation of educational materials using large language models (LLMs) is becoming increasingly common, but assigning difficulty levels to such materials still requires substantial human effort. LLM-as-a-Judge has therefore attracted attention, yet disagreement with human raters remains a major challenge. We propose a method for predicting which LLM-generated difficulty ratings are likely to disagree with human raters, so that such cases can be sent for re-rating.
Unlike prior approaches, our method does not rely on generation-time probability signals, which must be collected during rating generation and are often difficult to compare across LLMs. Instead, exploiting the fact that difficulty is an ordinal scale, we use a separate embedding space, such as ModernBERT, and identify disagreement candidates based on the geometric consistency of the rating set.
Experiments on English CEFR-based sentence difficulty assessment with GPT-OSS-120B and Qwen3-235B-A22B showed that the proposed method achieved higher AUC for predicting disagreement with human raters than probability-based baselines.

\end{abstract}


\section{Introduction}
In recent years, improvements in the performance of large language models (LLMs) and the development of supporting infrastructure have made it increasingly practical to automatically generate educational content such as practice questions, example sentences, and explanations for learners. As a result, the potential for educational applications has been expanding rapidly. At the same time, whether instructional materials are at an \textit{appropriate difficulty level} for learners is directly related to both learning effectiveness and fairness. Difficulty assessment is therefore a foundational task in the deployment of generated educational materials. However, manual difficulty rating requires expertise and time, and thus easily becomes a bottleneck in large-scale operation.

To address this human evaluation cost, \textit{LLM-as-a-Judge}, in which an LLM also serves as an evaluator, has attracted increasing attention \cite{llmasajudge}. LLM-as-a-Judge can reduce the temporal and financial costs of evaluation by lowering the burden on teachers and other human evaluators who would otherwise need to assess large numbers of examples and instructional materials for difficulty and related properties. At the same time, however, reliability, robustness, and alignment with human judgments have repeatedly been identified as major challenges \citep{chen-etal-2024-humans,raina-etal-2024-llm,chiang-etal-2024-large}. Therefore, in practical use, it is important not to accept LLM ratings as they are, but instead to efficiently identify cases that are likely to disagree with human judgments, that is, candidate cases of inter-rater disagreement.

One promising cue for identifying such disagreement candidates is to use generation probabilities output by the LLM at rating time, such as log-likelihoods or probability distributions. However, this approach has practical limitations: such signals are available only at generation time, so if they are not collected then, they cannot be recovered later. In addition, because generation probabilities differ across models, they are difficult to compare across LLMs.

Building on this motivation, this study extends the framework proposed in  \cite{ehara2025icce} and proposes a method for predicting, without using LLM probability signals, which difficulty ratings produced by LLM-as-a-Judge are likely to disagree with human raters. The key idea of this study is that difficulty is an ordinal scale.
In language difficulty estimation and readability assessment, deep representations such as BERT embeddings have been shown to be effective \citep{deutsch-etal-2020-linguistic}, and it is also well known that embedding spaces exhibit geometric structure, including directionality and anisotropy \citep{ethayarajh-2019-contextual}. Based on this observation, we prepare, independently of the LLM used as the rater, a separate embedding space such as that of ModernBERT \citep{warner2024modernbert}, and detect as disagreement candidates those ratings that are geometrically inconsistent with the ordinal difficulty structure, that is, with the alignment and consistency of the rating set as a whole.

This approach has two major practical advantages. First, because it does not depend on probability outputs, it can be applied even when such signals were not collected at the time the LLM-as-a-Judge generated its ratings. Second, unlike probability outputs, the quantities used by our method are geometric, which makes them comparable across difficulty labels assigned by different LLMs on the same dataset. Furthermore, the model used for the embedding space is much smaller than the LLM used for rating, with fewer than 1 billion parameters.

In our experiments, we consider an English CEFR-based difficulty assessment task, which provides internationally shared and rigorous difficulty criteria, and predict disagreement with human ratings for sentence difficulty labels produced by the open-weight LLMs gpt-oss-120b \citep{openai2025gptoss} and Qwen3-235B-A22B \citep{yang2025qwen3}. The results show that the proposed method predicts disagreement with higher AUC than methods based on LLM probability signals, demonstrating its effectiveness as a probability-free and interpretable method for prioritizing cases for re-rating.

\section{Related Work}
Methods for LLM-as-a-Judge, which assess the extent to which LLM evaluations are correct, have often relied on LLM probability signals.
For example, LLM-Rubric presents a framework that calibrates human ratings by using the probability distribution over rubric-based questions as input \citep{hashemi-etal-2024-llm}. However, the use of probability signals still poses practical challenges: (i) they must be obtained at the same time as the rating is generated, and are therefore affected by logging availability and API constraints; and (ii) probability scales and calibration differ across models, making comparison and interpretation difficult. In addition, as a black-box approach to uncertainty estimation that does not rely on probability distributions, previous work has proposed detecting anomalies from the self-consistency of sampled outputs (e.g., SelfCheckGPT) \citep{manakul-etal-2023-selfcheckgpt}, although such methods may require temperature adjustment or other sampling controls.

In language learning research, assigning difficulty levels based on a common framework is important, but it has also been pointed out that constructing sufficiently large, high-quality annotated datasets is itself difficult. For example, the study that constructed a CEFR-based sentence difficulty corpus (CEFR-SP) and demonstrated sentence difficulty estimation using BERT-based models was motivated by this challenge \citep{arase-etal-2022-cefr}.

\begin{figure}[t]
   \includegraphics[width=0.9\columnwidth]{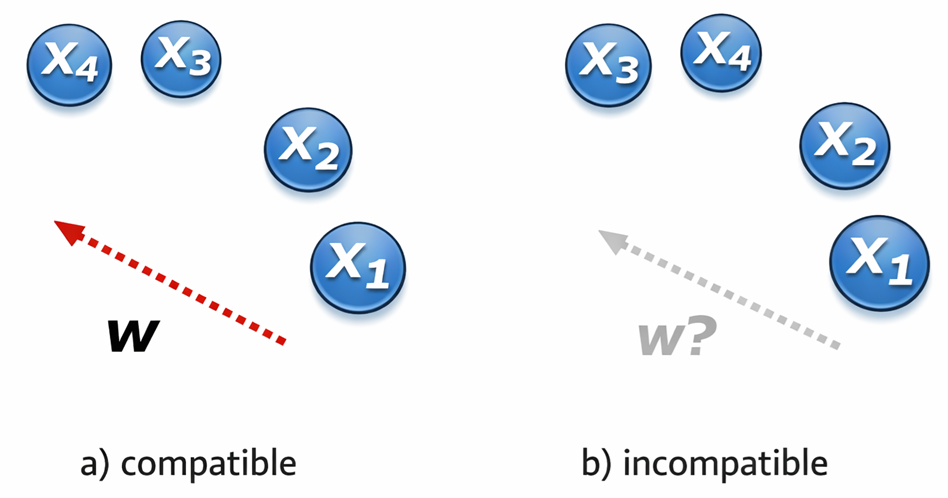}
   \caption{Figure reproduced from \cite{ehara2025icce} with their permission. In (a), let $\mathbf{x}_1, \mathbf{x}_2, \mathbf{x}_3$, and $\mathbf{x}_4$ denote two-dimensional word or sentence embeddings. We write $\mathbf{x}_1 \ord \mathbf{x}_2 \ord \mathbf{x}_3 \ord \mathbf{x}_4$ to indicate that $\mathbf{x}_1$ is annotated as easier than $\mathbf{x}_2$, $\mathbf{x}_2$ as easier than $\mathbf{x}_3$, and so on. In this case, if the points are ordered along the direction of $\mathbf{w}$ in the two-dimensional space, the resulting order matches the annotation. In contrast, in example (b), no such direction exists. In practice, most datasets and embedding spaces are closer to the situation in (b), but the degree of alignment is defined by how well one can find a direction that satisfies the ordering in (a) with as few constraint violations as possible. In addition, for a given instance, the degree to which the average of the difference vectors between that instance and all other instances is aligned with $\mathbf{w}$ is defined as the alignment score of that instance.}
\label{fig:intro}
\end{figure}

\section{Proposed Method}
The proposed method extends the framework proposed in \cite{ehara2025icce}. We first provide an intuitive explanation of the method using \figref{fig:intro}.
First, we prepare an embedding vector space that is \emph{completely separate} from the one used for difficulty labeling. For example, we assume a well-known encoder such as BERT \cite{devlin_bert_2019} or ModernBERT \cite{warner2024modernbert}, and all embedding vectors are assumed to be normalized to unit length.
The main idea of \cite{ehara2025icce} is that, given such an embedding space together with a set of difficulty labels assigned by an LLM, one can uniquely compute a vector that represents the degree of geometric consistency between the embedding space and the difficulty labeling.

While \cite{ehara2025icce} evaluates only the consistency of the difficulty labeling as a whole, it is also possible to compute the degree of consistency between the embedding space and each individual difficulty label. The proposed method uses this property: difficulty labels with low consistency with the embedding space are treated as candidates for low inter-rater agreement, that is, cases likely to show disagreement when another rater is introduced.
We now describe their method in more detail.
Consider $N$ embedding vectors $\{\mathbf{x}_1, \ldots, \mathbf{x}_N\}$, where each $\mathbf{x}_i$ is a $D$-dimensional vector.
We assume that all embedding vectors are normalized, that is, $||\mathbf{x}_i|| = 1$, where $||\mathbf{x}_i||$ denotes the Euclidean norm of the vector.
Let $\mathbf{w} \in \mathbb{R}^D$ be a $D$-dimensional vector representing a direction in the embedding space.
They convert difficulty labels into pairwise constraints between two instances. They denote by $\mathbf{x}_i$ the vector judged to be \emph{easier}, and by $\mathbf{x}_j$ the vector judged to be \emph{more difficult}. Then, the condition that instances are aligned in the embedding space in order of difficulty when viewed along the direction $\mathbf{w}$ can be written as $\mathbf{w}^\top \left(\mathbf{x}_i - \mathbf{x}_j\right) < 0.$
To transform the inequality constraints into equality constraints, we introduce variables that represent the degree to which each constraint is satisfied (slack variables), and write $\mathbf{w}^\top \left(\mathbf{x}_{i_k} - \mathbf{x}_{j_k}\right) + \xi_k = 0.$
Here, $k$ is an index over constraints.

At this point, by taking into account that $\mathbf{w}$ is a direction vector, that is, $||\mathbf{w}||^2 = 1$, and searching for the direction $\mathbf{w}$ that maximizes the sum of all $\xi_k$ values (the degrees to which the constraints are satisfied), we obtain the following mathematical optimization problem. Here, although originally $T = 0$, if we relax it to $T = -\infty$, the optimal $\mathbf{w}$ can be computed in closed form simply as the mean vector of the differences between labeled pairs.
\begin{equation}
    \max_{\mathbf{w}, \mathbf{\xi}} \sum_{k=1}^{K} \xi_k,
\end{equation}
\begin{equation}
    \textrm{s.t. } \forall k, \quad \mathbf{w}^\top (\mathbf{x}_{i_k} - \mathbf{x}_{j_k}) + \xi_k = 0, \quad \xi_k \geq T, \; ||\mathbf{w}||^2 = 1 \label{eq:jminusi}
\end{equation}

The solution to \eqref{eq:jminusi} is given by $\mathbf{w} \propto - \sum_{k=1}^K (\mathbf{x}_{i_k} - \mathbf{x}_{j_k})$.
Let $\bar{\mathbf{x}}_i$ denote the mean vector of level $i$, and similarly for level $j$, and let $N_i$ and $N_j$ denote the numbers of instances at levels $i$ and $j$, respectively. Then, $\mathbf{w}$ can be written as $\mathbf{w} \propto -(N_i \bar{\mathbf{x}}_i - N_j \bar{\mathbf{x}}_j)$. We refer to the method using this direction as ``exactw.''
Although the exact solution can be written in this form, in practice, when the numbers of instances at some levels are small as a result of LLM-based difficulty labeling, computing $\mathbf{w}$ exactly may lead to a biased direction vector.
Since $\mathbf{w}$ only needs to be a direction vector, we also propose a simpler method, ``most2,'' in which $\mathbf{w}$ is defined as the difference between the mean vectors of the two labels that were assigned most frequently by the LLM.

Once the direction $\mathbf{w}$ has been obtained, the values of $\xi_k$ can be computed accordingly. Then, by summing the $\xi_k$ values over all constraints involving item $i$, we can compute the degree of consistency for item $i$.

\section{Experimental Setup}
For the evaluation in this study, a dataset with rigorously obtained human ratings is desirable, and it is also important that the dataset include difficulty labels provided by multiple human raters. For this reason, we used the CEFR-SP dataset \cite{arase-etal-2022-cefr}. In this dataset, the difficulty of English sentences is rated by two expert annotators on a six-level scale, A1, A2, B1, B2, C1, and C2, based on the CEFR (Common European Framework of Reference for Languages), which is a standard and widely used framework for English proficiency in English language education.

Next, in this study, we needed baseline models from which confidence scores (token generation probabilities) could be obtained.
For this purpose, we used GPT-OSS-120B and Qwen3-235B-A22B (hereafter, Qwen3) for labeling, as they were among the largest models that could be run locally on a single GPU as of February 2026.

The input consisted of 10,004 English sentences. The label distribution produced by GPT-OSS-120B was A1: 500, A2: 3,345, B1: 2,899, B2: 3,230, C1: 29, and C2: 1. For Qwen3, the distribution was A1: 635, A2: 1,626, B1: 4,014, B2: 3,293, and C1: 436.

Performance in disagreement detection was evaluated using ROC-AUC, PR-AUC, and Precision@20. ROC-AUC is the area under the ROC curve obtained by thresholding the decision function and plotting the resulting pairs of true positive rate and false positive rate. PR-AUC is the area under the precision--recall curve obtained by similarly sweeping the threshold. Precision@20 denotes the precision when the top 20 instances ranked in descending order of $s(i)$ are regarded as candidates for re-evaluation.
For all metrics, sentences for which the label assigned by the LLM differed from that of human rater A or B were treated as positive instances.

\subsection{Difficulty Labeling by LLMs}
For the logprob-based baseline, we used the token log-probability of the label sequence output by GPT-OSS-120B or Qwen3 (logprob\_label) as the confidence score for disagreement detection. \par
The prompt included the CEFR criteria descriptions in the user input, and in practice we used the following System/User strings exactly as written:
\begin{quote}
System: You are an English proficiency rater who must assign a CEFR level (A1, A2, B1, B2, C1, or C2) to a single sentence using the provided criteria.
\end{quote}

After this, we directly inserted the English descriptions of the CEFR criteria for each level as given at \url{https://en.wikipedia.org/wiki/Common_European_Framework_of_Reference_for_Languages}.

For the user input, we used the following prompt:
\begin{quote}
You must output exactly one label token (A1, A2, B1, B2, C1, or C2) with no explanation.

User: Sentence: "{sentence}"
Respond with exactly one of: A1, A2, B1, B2, C1, C2.

Assistant:
\end{quote}
For Qwen3 instruct models, we additionally appended ``\texttt{\textbackslash nDo not output anything else. Answer: }'' at the end in order to constrain the response to a single label only.

\subsection{Embedding Models Used}
The alignment measure of \cite{ehara2025icce} estimates a single direction $w$ using the differences between class centroids for each CEFR level, and measures the degree of constraint violation by the inner product with $w$. In this study, we implemented the aforementioned ``exactw'' as well as ``most2,'' which uses the difference between the centroids of the two most frequent labels.
The embedding models used for comparison are shown in \tabref{tab:model_params}. Except for ModernBERT, the names correspond to the model names on Hugging Face (\url{https://huggingface.co/}).

\begin{table}[t]
  \centering
  \caption{Number of parameters in the embedding models} \label{tab:model_params}
  \scalebox{0.7}{
  \begin{tabular}{lc}
    \hline
    Model & Number of parameters (M) \\ 
    \hline
    \texttt{modernbert} & 394.8 \\
    \texttt{intfloat\slash e5-large-v2} & 335.1 \\ 
    \texttt{BAAI\slash bge-large-en-v1.5} & 335.1 \\ 
    \texttt{intfloat\slash multilingual-e5-large-instruct} & 559.9 \\ 
    \texttt{sentence-transformers\slash LaBSE} & 470.9 \\ 
    \texttt{microsoft\slash deberta-v3-large} & 434.0 \\ 
    \texttt{sentence-transformers\slash sentence-t5-large} & 737.7 \\ 
    \texttt{sentence-transformers\slash sentence-t5-base} & 222.9 \\ 
    \texttt{sentence-transformers\slash all-roberta-large-v1} & 355.4 \\ 
    \texttt{Alibaba-NLP\slash gte-large-en-v1.5} & 434.1 \\ 
    \texttt{hkunlp\slash instructor-large} & 737.7 \\ 
    \hline
  \end{tabular}}
\end{table}

\section{Experimental Results}
Let us first restate the goal of this study.
We first assign difficulty labels using an LLM. We then use various methods to identify, from those difficulty labels, the cases that are likely to disagree with human raters.
Because the CEFR-SP dataset used in this study contains two human raters, we conducted experiments for both disagreement with Rater A and disagreement with Rater B.
As noted above, for all three metrics---ROC-AUC, PR-AUC, and Precision@20---higher values indicate better performance in predicting disagreement.
\tabref{tab:gptossA} and \tabref{tab:gptossB} show the results when GPT-OSS-120B was used for difficulty labeling, while \tabref{tab:qwen3A} and \tabref{tab:qwen3B} show the results when Qwen3 was used for difficulty labeling.
Methods with \emph{logprob} in their names are the conventional baselines that use the token generation probabilities of each LLM's output labels.
Since no previous method has proposed using embedding vectors separate from the LLM itself, all other methods are proposed methods.

The \emph{exactw} method uses, as the direction of $\mathbf{w}$, the average of all pairwise constraints.
In contrast, \emph{most2} uses as $\mathbf{w}$ the direction of the difference between the mean vectors of the two levels that were assigned most frequently and second most frequently by the LLM.

First, across all four combinations of GPT-OSS-120B or Qwen3 with Rater A or B, modernbert (most2) achieved higher PR-AUC than the conventional logprob-based method.
Previously, the only option was to use the confidence scores output by the LLM at the time of response generation, and once these were not collected during generation, it was difficult to obtain them again afterward.
In contrast, the proposed method makes it possible to obtain, after the response has already been generated, a more effective indicator of confidence for LLM-as-a-Judge.
This is a novel and practically useful result.

Second, \emph{for most embedding models, including modernbert, most2 performed better than exactw}.
This is an important finding.
A likely reason is that the difficulty level labels assigned by each LLM are not evenly distributed.
Rather than measuring consistency with the embedding space while also taking into account instances assigned to infrequent levels, better performance was obtained, from the perspective of detecting the magnitude of disagreement with human raters, by closely examining the levels most frequently assigned by the LLM and measuring their consistency with the embedding space.

Furthermore, although various embeddings were tested, \emph{overall, modernbert showed strong performance}.
ModernBERT \cite{warner2024modernbert} is a method that comprehensively improves upon BERT, and like BERT, it is designed not for a specific dataset but for broad use of embedding vectors across a wide range of tasks.
The proposed method in this study uses modernbert in a way that is consistent with its original design intention.
It is therefore reasonable to interpret the result as showing that, even with a relatively small number of parameters, an embedding model used in accordance with its intended design is the most useful.

\begin{table}[t]
  \centering
  \caption{Evaluation metrics for GPT-OSS-120B (Rater A)}
  \label{tab:gptossA}\scalebox{0.55}{
  \begin{tabular}{lccc}
    \hline
    Method (Rater) & ROC-AUC & PR-AUC & precision@20 \\
    \hline
        logprob\_label (A) & 0.561 & 0.579 & 0.80 \\
    p\_i exactw / chain (A) & 0.509 & 0.528 & 0.55 \\
    modernbert \cite{ehara2025icce} most2  & 0.606 & 0.652 & 1.00 \\
    modernbert \cite{ehara2025icce} exactw  & 0.424 & 0.503 & 0.80 \\\hline\hline
    \texttt{e5-large-v2} & 0.578 & 0.624 & 1.00 \\ 
    \texttt{bge-large-en-v1.5} & 0.575 & 0.614 & 1.00 \\ 
    \texttt{multilingual-e5-large-instruct} & 0.602 & 0.652 & 1.00 \\ 
    \texttt{LaBSE} & 0.588 & 0.637 & 1.00 \\ 
    \texttt{deberta-v3-large} & 0.589 & 0.622 & 0.95 \\ 
    \texttt{sentence-t5-large} & 0.568 & 0.604 & 1.00 \\ 
    \texttt{sentence-t5-base} & 0.571 & 0.610 & 1.00 \\ 
    \texttt{all-roberta-large-v1} & 0.575 & 0.608 & 0.95 \\ 
    \texttt{gte-large-en-v1.5} & 0.574 & 0.611 & 0.95 \\ 
    \texttt{instructor-large} & 0.577 & 0.618 & 1.00 \\ 

    \hline
  \end{tabular}}
\end{table}

\begin{table}[t]
  \centering
  \caption{Evaluation metrics for GPT-OSS-120B (Rater B)}
  \label{tab:gptossB}\scalebox{0.55}{
  \begin{tabular}{lccc}
    \hline
    Method (Rater) & ROC-AUC & PR-AUC & precision@20 \\
    \hline
    logprob\_label (B) & 0.558 & 0.668 & 0.95 \\
    p\_i exactw / chain (B) & 0.511 & 0.653 & 0.85 \\
    modernbert \cite{ehara2025icce} most2 / chain (B) & 0.547 & 0.690 & 1.00 \\
    modernbert \cite{ehara2025icce} exactw / chain (B) & 0.427 & 0.578 & 0.70 \\ \hline\hline
    \texttt{e5-large-v2} & 0.552 & 0.676 & 1.00 \\ 
    \texttt{bge-large-en-v1.5} & 0.535 & 0.666 & 1.00 \\ 
    \texttt{multilingual-e5-large-instruct} & 0.548 & 0.683 & 1.00 \\ 
    \texttt{LaBSE} & 0.551 & 0.685 & 1.00 \\ 
    \texttt{deberta-v3-large} & 0.540 & 0.675 & 1.00 \\ 
    \texttt{sentence-t5-large} & 0.530 & 0.665 & 1.00 \\ 
    \texttt{sentence-t5-base} & 0.531 & 0.665 & 1.00 \\ 
    \texttt{all-roberta-large-v1} & 0.539 & 0.661 & 1.00 \\ 
    \texttt{gte-large-en-v1.5} & 0.535 & 0.665 & 0.95 \\ 
    \texttt{instructor-large} & 0.530 & 0.668 & 1.00 \\ 

    \hline
  \end{tabular}}
\end{table}

\begin{table}[t]
  \centering
  \caption{Evaluation metrics for Qwen3 (Rater A)}
  \label{tab:qwen3A}\scalebox{0.55}{
  \begin{tabular}{lccc}
    \hline
    Method (Rater) & ROC-AUC & PR-AUC & precision@20 \\
    \hline
logprob\_label (A) & 0.567 & 0.539 & 0.55 \\
p\_i exactw / chain (A) & 0.525 & 0.508 & 0.60 \\
\cite{ehara2025icce} most2 / chain (A) & 0.593 & 0.591 & 0.90 \\
\cite{ehara2025icce} exactw / chain (A) & 0.482 & 0.509 & 0.75 \\\hline\hline
 \texttt{e5-large-v2} & 0.534 & 0.537 & 0.80 \\ 
    \texttt{bge-large-en-v1.5} & 0.540 & 0.538 & 0.75 \\ 
    \texttt{multilingual-e5-large-instruct} & 0.567 & 0.568 & 0.95 \\ 
    \texttt{LaBSE} & 0.545 & 0.544 & 0.95 \\ 
    \texttt{deberta-v3-large} & 0.565 & 0.551 & 0.80 \\ 
    \texttt{sentence-t5-large} & 0.538 & 0.532 & 0.55 \\ 
    \texttt{sentence-t5-base} & 0.543 & 0.537 & 0.60 \\ 
    \texttt{all-roberta-large-v1} & 0.544 & 0.536 & 0.75 \\ 
    \texttt{gte-large-en-v1.5} & 0.540 & 0.538 & 0.80 \\ 
    \texttt{instructor-large} & 0.542 & 0.539 & 0.85 \\ 
    \hline
  \end{tabular}}
\end{table}

\begin{table}[t!]
  \centering
  \caption{Evaluation metrics for Qwen3 (Rater B)}
  \label{tab:qwen3B}\scalebox{0.55}{
  \begin{tabular}{lccc}
    \hline
    Method (Rater) & ROC-AUC & PR-AUC & precision@20 \\
    \hline
logprob\_label (B) & 0.539 & 0.586 & 0.85 \\
p\_i exactw / chain (B) & 0.501 & 0.558 & 0.75 \\
\cite{ehara2025icce} most2 / chain (B) & 0.537 & 0.610 & 0.85 \\
\cite{ehara2025icce} exactw / chain (B) & 0.424 & 0.508 & 0.30 \\\hline\hline
    \texttt{e5-large-v2} & 0.520 & 0.581 & 0.70 \\ 
    \texttt{bge-large-en-v1.5} & 0.506 & 0.577 & 0.70 \\ 
    \texttt{multilingual-e5-large-instruct} & 0.526 & 0.592 & 0.75 \\ 
    \texttt{LaBSE} & 0.533 & 0.584 & 0.75 \\ 
    \texttt{deberta-v3-large} & 0.523 & 0.585 & 0.75 \\ 
    \texttt{sentence-t5-large} & 0.503 & 0.569 & 0.75 \\ 
    \texttt{sentence-t5-base} & 0.496 & 0.567 & 0.80 \\ 
    \texttt{all-roberta-large-v1} & 0.503 & 0.570 & 0.70 \\ 
    \texttt{gte-large-en-v1.5} & 0.505 & 0.574 & 0.75 \\ 
    \texttt{instructor-large} & 0.503 & 0.573 & 0.80 \\ 
    \hline
  \end{tabular}}
\end{table}

\section{Conclusion}
In this study, we proposed a reliability indicator for individual difficulty labels assigned by LLM-as-a-Judge that is easier to obtain and more effective than conventional probability-based approaches. Unlike prior methods, the proposed method requires only the LLM-assigned labels and the original text, without relying on generation-time confidence scores.
The method exploits the fact that difficulty is an ordinal scale and measures, for each label, the geometric consistency between an additional embedding space and the difficulty labels assigned by the LLM. Experiments on CEFR-SP showed that the proposed method can detect labels likely to disagree with human raters more accurately than conventional methods. It is also memory-efficient, since it uses only text embedding models with fewer than 1B parameters.

As future work, although this study was evaluated on English text difficulty assessment, extending the method to STEM subjects will require establishing a rigorous difficulty scale comparable to CEFR across human raters. If such a scale can be developed, the proposed method may also be applied to STEM difficulty assessment.

\appendix
\section{Scope and Limitations}

The proposed method should be understood as a method for
detecting potentially unreliable difficulty labels in settings
where the target difficulty construct is ordinal and is at least
partly reflected in an independent representation space. In this
paper, we evaluated the method on CEFR-based English sentence
difficulty assessment. This is a favorable setting for the proposed
approach because the target construct is linguistic difficulty, and
text embedding models such as ModernBERT are expected to encode
lexical, syntactic, and semantic properties that are relevant to
language proficiency levels.

This scope is important. The proposed method is not intended to be
a domain-independent solution to all forms of educational difficulty
assessment. In particular, difficulty in STEM domains may depend on
conceptual prerequisites, problem-solving procedures, mathematical
structure, required operations, common misconceptions, and the
interaction between textual and non-textual representations. These
factors may not be adequately captured by standard text embeddings.
In such cases, geometric consistency in a generic embedding space may
reflect lexical or semantic similarity rather than the educational
difficulty of the item.

Therefore, applying the proposed method to STEM difficulty assessment
would require additional domain-specific assumptions and validation.
One possible direction is to replace the generic text embedding space
with a representation that incorporates prerequisite relations,
knowledge-component annotations, worked-solution structure, or other
features that are known to be relevant to the target domain. Another
requirement is a reliable ordinal difficulty scale with sufficient
agreement among human raters, comparable in role to CEFR in the
present study.

Accordingly, the main contribution of this paper is not that embedding
geometry universally determines educational difficulty. Rather, the
contribution is to show that, in language-based difficulty assessment,
an independently obtained embedding space can provide a practical
and probability-free signal for prioritizing LLM-generated ratings
that are likely to disagree with human raters. Extending this idea
beyond language difficulty assessment is an important direction for
future work, but it should be treated as a domain-specific research
problem rather than as a direct application of the present method.

\section*{Acknowledgments}

This work was supported by JST PRESTO, Japan, Grant Number JPMJPR2363 and by JSPS KAKENHI Grant Number JP22K12287.
We are deeply grateful to the anonymous reviewers for their constructive feedback.
%
\bibliographystyle{abbrv}
\bibliography{_custom}
\balancecolumns
\end{document}